\documentclass[letterpaper]{article}

\usepackage{natbib,alifeconf,float, amsmath}  
\title{Neuroevolving Electronic Dynamical Networks}

\author{Derek Whitley$^{1,2}$ \\ \\
$^1$Vivum AI, 642 N Madison Ave, Bloomington, IN 47404 \\
$^2$Indiana University, Bloomington, IN 47405 \\
derek@vivum.ai \\
\mbox{}\\
}

\begin{document}
\maketitle

\begin{abstract}

Neuroevolution is a powerful method of applying an evolutionary algorithm to refine the performance of artificial neural networks through natural selection; however, the fitness evaluation of these networks can be time-consuming and computationally expensive, particularly for continuous time recurrent neural networks (CTRNNs) that necessitate the simulation of differential equations. To overcome this challenge, methods from evolvable hardware (EHW) offer a solution. The application of evolutionary algorithms to modern system-on-chip (SoC) field programmable gate arrays (FPGAs) that have the ability to undergo dynamic and partial reconfiguration and communicate bilaterally through high speed interfaces (e.g. AXI) enables the extremely rapid evaluation of the fitness of CTRNNs, effectively addressing the update and evaluation bottlenecks associated with conventional methods of evolvable hardware. By incorporating fitness evaluation directly upon the programmable logic of the FPGA, hyper-parallel evaluation becomes feasible, dramatically reducing the time required for assessment. This inherent parallelism of FPGAs accelerates the entire neuroevolutionary process by several orders of magnitude, facilitating faster convergence to an optimal solution. The work presented in this study demonstrates the potential of utilizing FPGAs as a powerful platform for neuroevolving dynamic neural networks.

\end{abstract}

\section{Introduction}

The neuroevolution of Continuous Time Recurrent Neural Networks (CTRNNs) \citep{hopfield1982neural,Funahashi1993OriginalCA, beer1995dynamics} is a time-intensive process of evaluating the fitness of time-integrating neural networks, which involves simulating differential equations that govern their dynamics. For non-dynamic neural models, such as convolutional neural networks \citep{lecun1998gradient, krizhevsky2012imagenet}, recurrent neural networks \citep{bengio1994learning}, long short-term memory networks \citep{hochreiter1997long}, and many other deep learning methods \citep{hinton2006fast,ng2011sparse}, differential calculations are unnecessary during the inference phase and avoided where possible during training, whereas they are required for CTRNNs for both training and inference. In contrast to typical deep learning methods, dynamic neural models such as CTRNNs do not have immediate out-of-the-box support from popular libraries like TensorFlow, PyTorch, or Keras without specific customization, nor do any of these libraries provide hardware acceleration such as general purpose graphical processing units (GPGPUs/GPUs) to dynamic models.

GPGPUs are a type of vector processor currently standard for training  most types of deep learning models due to their ability to perform massive parallel computations, which are well-suited for deep learning's large volume matrix operations, resulting in faster training and inference times than traditional CPUs. However, FPGAs can offer better speed and performance-to-power ratio than GPUs for certain tasks, particularly those that require time-integration. This is because FPGAs can be dynamically reconfigured at the architectural level to execute specific computations efficiently. This ability is critically valuable in tasks that involve expressing complex time-dependent behaviors, such as the simulation of dynamical systems or multivariate control tasks, granting them a competitive edge over GPUs and CPUs, especially in scenarios where energy efficiency is of paramount importance. As the industry evolves, FPGAs are poised to become an increasingly attractive alternative to GPUs, capable of reshaping the landscape of dynamic and liquid artificial neural network training and implementation.

As FPGAs continue to gain traction, there is a growing interest in harnessing their capabilities for neuroevolution for the following reasons \citep{floreano2008neuroevolution}:
\begin{itemize}
\item \textbf{Global optimization:} Neuroevolution is an evolutionary algorithm that employs search and optimization techniques inspired by natural selection. This approach enables the exploration of a broader solution space, making it more likely to discover global optima. Gradient descent, on the other hand, is a local optimization method and can often get trapped in local minima, failing to reach the global minimum.
\item \textbf{Complex fitness landscapes:} Neuroevolution is particularly suitable for addressing problems with complex and irregular fitness landscapes, where gradient-based methods might struggle. Evolutionary algorithms are inherently robust and can handle noise, discontinuities, and non differentiability in the fitness function, which may be challenging for gradient descent \citep{amari1993backpropagation}.
\item \textbf{Topology and weight optimization:} Neuroevolution algorithms can optimize both the architecture (topology) and the weights of neural networks simultaneously. In contrast, gradient descent primarily focuses on adjusting the weights of a predetermined network structure. The ability to optimize network topology allows neuroevolution to discover more efficient and effective network configurations.
\item \textbf{Parallelism:} Neuroevolution algorithms can naturally take advantage of the intrinsic parallelism of FPGAs. The evaluation of multiple candidate solutions can be performed concurrently, accelerating the evolutionary process. Gradient descent, while benefiting from some parallelism, is typically limited by the sequential nature of weight updates.
\item \textbf{Customizability and adaptability:} FPGAs provide a reconfigurable hardware platform, allowing for the design and implementation of custom circuits for specific computational tasks. This flexibility enables FPGAs to achieve high performance and low latency, making them particularly suitable for real-time applications and edge computing. In contrast, gradient descent on traditional platforms might not offer the same level of customization and adaptability.
\end{itemize}

By integrating the fitness evaluation algorithm into the programmable logic of the FPGA, parallel evaluation becomes feasible, substantially reducing the time required for assessment. In an effort to unlock each of these benefits, this research explores the use of FPGAs as a potent platform for neuroevolving dynamic neural networks, illustrating the potential to overcome the limitations posed by conventional CPU-based evaluation techniques. 

\subsection{FPGAs}
Field Programmable Gate Arrays (FPGAs) \citep{guo2017survey} are integrated circuits that can be programmed and reprogrammed by users to perform a wide range of custom-designed computations. Unlike traditional CPUs and GPUs, which have fixed architectures that limit their ability to perform certain types of computations efficiently, FPGAs can be customized to meet the specific requirements of a particular application. This adaptability makes FPGAs a highly flexible and versatile computational platform, with potential applications in fields ranging from telecommunications to image processing to artificial intelligence.

FPGAs consist of a large array of configurable logic blocks that can be interconnected to form custom-designed circuits. These logic blocks can be configured to perform different types of logical operations, such as AND, OR, and NOT, and can be combined to create complex circuits that can implement algorithms and mathematical functions. FPGAs also typically include on-chip memory blocks and specialized blocks for performing tasks such as digital signal processing and floating-point arithmetic. Because FPGAs can be programmed at the hardware level, they can be designed to perform computations with low latency and high throughput, making them a popular choice for applications that require real-time processing or low power consumption.

In recent years, FPGAs have gained popularity for implementing myriad types of artificial neural networks (ANNs) due to their high performance and low power consumption. Traditionally, ANNs are implemented on general-purpose computing platforms, such as CPUs or GPUs, using software frameworks such as TensorFlow, Keras, or PyTorch. However, as ANNs become larger and more complex, the computational demands of training and running them increase, making them impractical for many applications. FPGAs, on the other hand, offer several advantages for implementing ANNs:

\begin{enumerate}
    \item FPGAs can be programmed to execute ANNs directly in hardware, which can result in significant performance improvements compared to software-based implementations. This is because FPGAs can perform per-neuron computations in parallel, unlike CPUs and GPUs, which have a limited number of processing cores. In addition, FPGAs can be customized to execute specific operations that are commonly used in ANNs, such as time-dependent  dot-product, matrix multiplication, or convolution, further enhancing their performance.
    
    \item They offer low power consumption, which is particularly important for applications that require ANNs to operate in low-power environments or with limited battery life. This is because FPGAs are designed to allow all components to operate asynchronously and with a low voltage supply, resulting in lower power consumption compared to CPUs and GPUs.

    \item FPGAs provide flexibility and reconfigurability, which can be particularly useful during the development and testing of ANNs. This is because they can be reconfigured to implement different ANN architectures and parameters, which can enable faster experimentation and prototyping. Moreover, FPGAs can be reconfigured in real-time, allowing for the implementation of hardware-accelerated ANNs that can adapt to changing conditions.
\end{enumerate}

In light of these advantages, however, one of the main challenges is the design of the ANN architecture and optimization of the FPGA implementation for efficient resource utilization. This requires a deep understanding of both ANNs and FPGAs. Moreover, the programming of FPGAs requires specialized skills and tools, which can be a barrier for some researchers and developers.

Despite these challenges, the use of FPGAs for implementing ANNs is becoming increasingly popular, with several companies offering FPGA-based solutions for ANN implementation, and with FPGA vendors offering dedicated hardware and software tools for FPGA-based ANN development. The ability of FPGAs to offer high performance, low power consumption, and flexibility make them an attractive platform for ANNs in a wide range of applications, from embedded systems to data centers \citep{boutros2020beyond}.

\subsection{CTRNNs}

Continuous time recurrent neural networks (CTRNNs) are a class of neural networks that are particularly well-suited for modeling dynamical systems \citep{Funahashi1993OriginalCA}. CTRNNs are able to capture the temporal evolution of a system's state, making them ideal for applications such as control, prediction, pattern recognition and generation \citep{fernando2003pattern}.

At their core, CTRNNs consist of a set of interconnected neurons that evolve through time according to a system of ordinary differential equations. The state of each neuron is represented by a continuous-valued variable, and the connections between neurons are described by a set of weights. The overall dynamics of the network are governed by the interaction between these continuous parameters and the activity of the neurons themselves. The model is traditionally written in the following form:

\begin{align}
\tau_i \frac{dy_i(t)}{dt} &= -y_i(t) + \sum_{j=1}^{N} w_{ij} \sigma(y_j(t) + \theta_j) + I_i(t), \\
\sigma(x) &= \frac{1}{1 + e^{-x}},
\end{align}

where:
$y_i(t)$ represents the state of the $i$-th neuron at time $t$,
$\tau_i$ is the time constant of the $i$-th neuron,
$w_{ij}$ is the weight of the connection from neuron $j$ to neuron $i$,
$\theta_j$ is the bias of neuron $j$,
$N$ is the total number of neurons in the network,
$I_i(t)$ is the external input to the $i$-th neuron at time $t$, and
$\sigma(x)$ is the sigmoid activation function.

In these equations, $x_{i}$ represents the state of neuron $i$, and $w_{ij}$ represents the weight of the connection from neuron $j$ to neuron $i$. The inputs $I_{i}$ represent any external inputs to the neurons, and $\sigma$ is an activation function (in this case, the standard sigmoid) that determines the output of the neuron. The differential equations describe how the state of each neuron changes over time, based on its current state, the inputs it receives, and the weights of its connections to other neurons.

Note that these equations can be modified to include additional neurons, inputs, and connections, depending on the specific requirements of the CTRNN model. Additionally, there are different types of activation functions that can be used, such as the sigmoid function or the hyperbolic tangent function, which can be chosen based on the specific requirements of the model.

When compared to most other deep learning models, the key advantage of CTRNNs is their ability to exhibit complex temporal behaviors. Unlike many types of deep learning networks, particularly feedfoward architectures which operate on a fixed input-output mapping, CTRNNs can generate rich and dynamic output patterns that evolve over time. This makes them particularly useful for time series tasks such as speech recognition, gesture recognition, and complex pattern generation, where the time dependent nature of the task domain is integral to the output behavior of the network. Another strength of CTRNNs is their ability to adapt to new environments. By dynamically adjusting the weights of the network, CTRNNs can optimize their performance for a given task. This makes them ideal for applications such as sensing, kinematics, and actuation, where the ability to adapt an output behavior to the environment in real-time is imperative.

Despite their many advantages, CTRNNs can be challenging to design and train. The complex dynamics of the network can lead to issues such as unstable behavior and slow convergence during training. Additionally, the computational cost of evaluating CTRNNs can be high, particularly for large-scale networks with many neurons and connections.

To address these challenges, researchers have developed a range of optimization techniques for training CTRNNs. These include traditional gradient-based methods such as backpropagation through time (BPTT), as well as other advanced techniques such as evolutionary algorithms and reinforcement learning. These methods allow for the efficient optimization of CTRNNs for a variety of tasks, from prediction and classification to the generation of complex output patterns and control.

\subsection{Neuroevolution}
Neuroevolution is the method of applying an evolutionary algorithm to refine the parameters of artificial neural networks through an algorithmic process resembling natural selection \citep{ronald1994}. This approach is quite different from traditional training methods, such as backpropagation, which adjusts the weights of a neural network using gradient descent. Instead, neuroevolution leverages the principles of Darwinian natural selection, mutating and recombining the weights, architectures, or learning rules of neural networks \citep{stanley2002evolving}. The selected networks survive and reproduce based on their performance in a given task, effectively allowing evolution to explore vast network configurations and parameter spaces.

Although this method has demonstrated effectiveness in various domains, its primary weakness remains in the computational cost usually associated with the fitness evaluation stage (Xue et al., 2016). Evaluating the fitness of each candidate solution, particularly in the context of CTRNNs which involve the accurate simulation of continuous differential equations, often generates lengthy lead times. This requirement can significantly stretch the duration of evolution, especially when searching for optimal network configurations for complex tasks.

The current methods for evaluating the fitness of CTRNNs are limited in their ability to handle large networks or large population sizes. Traditional approaches rely on sequential evaluations, where each network is evaluated one at a time, which can take several minutes to hours depending on the complexity of the network. This not only increases the computational time but also limits the number of evaluations that can be performed in a reasonable amount of time, hindering the optimization process.

Moreover, evaluating the fitness of CTRNNs in parallel is not straightforward due to the complex nature of the evaluation process. Parallelizing the evaluation of CTRNNs requires careful consideration of the inter-dependencies between neurons and the synchronization of the simulation process. These factors make it challenging to exploit the full potential of parallel computing, further exacerbating the computational time required for fitness evaluation. Therefore, addressing the bottleneck in fitness evaluation for CTRNNs through neuroevolution is a critical challenge that needs to be addressed to realize the full potential of this optimization technique.

\subsection{Accelerating Neurovolution with FPGAs}
As previously described, CTRNNs are an effective class of artificial neural networks that are widely used in a variety of applications such as robotics, control systems, and signal processing. However, evaluating the fitness of CTRNNs trained via neuroevolution is time-consuming and computationally expensive due to the simulation of differential equations, limiting the scalability of neuroevolutionary approaches for optimizing CTRNNs.

Many modern FPGAs have high speed communication interfaces, such as the advanced extensible interface (AXI), and dedicated electronics that enable the modification of a specific portion of the FPGA hardware architecture while the remaining portion remains active (e.g. dynamic partial reconfiguration). This capability enables the rapid reconfiguration of the FPGA to perform specific tasks, allowing for parallelism and faster computation time. By implementing the fitness evaluation algorithm on the programmable logic of the FPGA, the evaluation can be performed in parallel, significantly reducing the evaluation time.

The proposed solution offers several advantages over traditional methods. First, the intrinsic parallelism of FPGAs can be leveraged to accelerate the time series fitness evaluation of CTRNNs, allowing for faster convergence to an optimal solution. Second, FPGAs enables the rapid deployment of CTRNNs to dedicated hardware, which improve the scalability of the neuroevolutionary approach. Third, the low power consumption of FPGAs compared to traditional processors can result in significant energy savings.

To implement the proposed solution, the computational representation of the CTRNN is directly programmed onto the FPGA as a circuit. The CTRNN is then emulated on the FPGA hardware, where the neural parameters are stored in the FPGA’s dedicated block memory. The fitness evaluation function is then performed in parallel by the FPGA for each individual in the population. The network outputs are captured by the host computer and used to calculate the fitness value to be used for the next generation of the neuroevolution process.

The proposed method has been shown to significantly reduce the evaluation time of CTRNNs, resulting in a faster convergence to an optimal solution. Additionally, the low power consumption of FPGAs can lead to significant energy savings in large-scale optimization applications.

\subsection{Objectives}

The purpose of this study is to determine the effectiveness of the previously proposed FPGA-based neuroevolution approach. It is important to evaluate its performance and efficiency in comparison to traditional methods, especially CPU-based neuroevolution. This comparative analysis considers various factors, including the time taken for fitness evaluations and the overall computational resources utilized. By demonstrating the superiority of FPGA-based neuroevolution in these aspects, compelling evidence for adopting this method over conventional alternatives can be derived.

Another critical aspect of the study is assessing the impact of FPGA-based neuroevolution on the convergence rate of the evolutionary process. Faster convergence to optimal or near-optimal solutions is highly desirable in real-world applications, as it enables more efficient problem-solving and potentially reduces the cost of experimentation. By quantifying the improvement in convergence speed achieved through FPGA-based neuroevolution, future researchers can benefit from this approach.

Scalability and adaptability are essential considerations in evaluating the merits of FPGA-based neuroevolution. To establish the robustness of the proposed method, this research explores its scalability in handling CTRNNs of increasing sizes.

Lastly, investigating the energy efficiency of the FPGA-based neuroevolution method is important, particularly in the context of power consumption and computational resources. As energy efficiency becomes an increasingly critical concern in modern computing, it is essential to demonstrate the advantages of FPGA-based neuroevolution over traditional alternatives in this regard. By quantifying the improvements in energy efficiency achieved through this approach, researchers can further strengthen the case for adopting FPGAs as a powerful platform for neuroevolving CTRNNs and other neural network architectures.

By addressing key objectives such as performance, efficiency, convergence rate, scalability, adaptability, and energy efficiency, a compelling case for the widespread adoption of this method becomes tangible in both academia and industry.

\section{Method}

In this study, a Xilinx ZCU102 evaluation board \citep{xilinx2020ultrascale} was used to embed a CTRNN onto the programmable logic FPGA and subsequently subjected to a differential analysis of the performance of the identical neuroevolutionary task performed on an ARM Cortex-A53 processor. This required several steps, including the design of the evaluation architecture, communication, and synchronization with the FPGA, as well as optimization of the implementation for performance and resource usage. 

The first step in this process was to design the evaluation architecture and determining maximum number of CTRNNs neurons that can be embedded in the logic fabric as possible. The evaluation architecture consists of multiple parallel processing elements (PEs), each capable of performing forward-pass simulations of the CTRNNs. These PEs are organized in such a way to ensure efficient data flow across the FPGA fabric. The configuration is designed in such a way to exploit parallelism, reducing the overall evaluation time for CTRNNs.

To implement the CTRNN model, each PE includes components for neuron state update, activation function computation, and weight matrix multiplication. For simplicity, neuron state updates are computed using forward Euler method for numerical integration, as well as a look-up table (LUT) containing 256 entries of fixed-point 16-bit elements describing the standard sigmoid function used to efficiently execute these computations. Overall, 628 neurons were able to fit on the ZCU102 without requiring further design optimization.

Synchronization between the FPGA and the host system is essential to ensure correct operation. This can be achieved using dedicated handshake control signals, which are responsible for initiating and terminating the evaluation process. Additionally, the host system can poll the FPGA to determine the completion status of the CTRNN fitness evaluation.

Communication between the host system and the FPGA is necessary for efficient fitness evaluation. The Xilinx ZCU102 FPGA features a high-speed AXI bus, which is leveraged to transfer the CTRNN model parameters, including neuron states, biases, connection weights, and time constants, between the FPGA and the onboard ARM processor. The AXI interface can also be used to read the output neuron states from the FPGA after the evaluation is completed. Efficient data transfer is ensured by using Direct Memory Access (DMA) engines, which offload the data movement tasks from the host CPU.

The experiment compared the performance of a single ARM core to the performance of an FPGA where each would run an identical copy of a neuroevolutionary algorithm designed to evolve a coupled oscillator. The desired output behavior of the coupled oscillator is such that the output of one neuron would drive the input of the other, and vice versa, generating a sinusoidal-like waveform. The only difference in experimental setup is that the fitness evaluation steps of the evolutionary algorithm were run on the programmable logic of the SoC-based FPGA while the ARM core was used to run the rest of the evolutionary algorithm.

\section{Experimental Results}

The primary focus for the results is on the total time required to evaluate an increasing number of CTRNNs for an increasing integration duration. The experiments were conducted using a range of evaluation period (from 100 to 1000, in increments of 100) to determine the performance of both implementations under varying workloads.

\subsection{FPGA versus ARM}

The FPGA is able to evaluate a large number of CTRNNs by capitalizing on the intrinsic parallelism of its logic fabric, reducing the overall time required to complete the evolutionary process. Furthermore, the high-bandwidth throughput of the AXI connection allows for the updating of the next generation's entire population of CTRNNs, which further reduces the time required to evaluate each generation. As a result, the FPGA implementation demonstrates a significant reduction in the total time required to complete the 1,000,000 total evaluations compared to the ARM processor-based implementation.

\begin{figure}[h]
\centering
\includegraphics[width=1.0\linewidth]{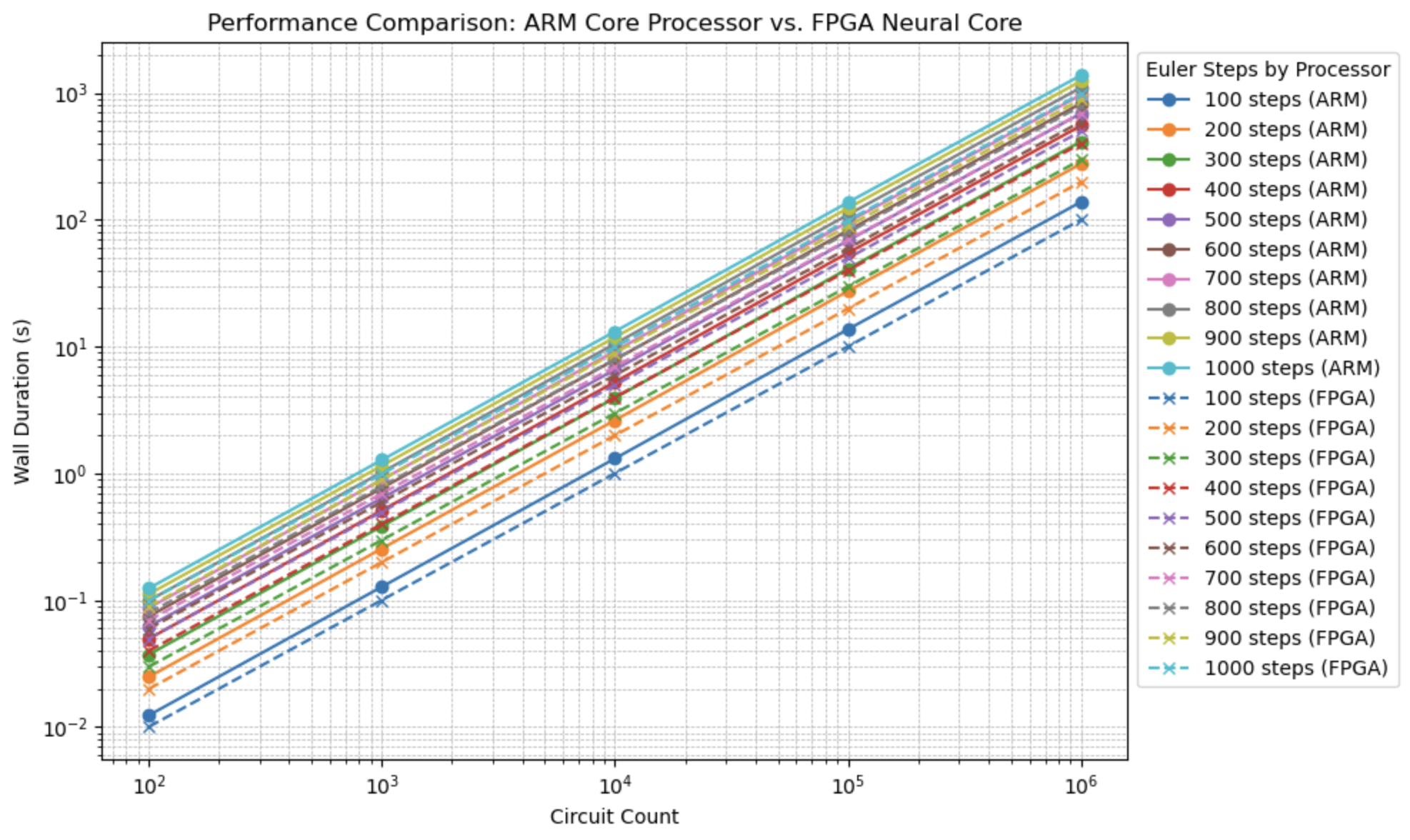}
\caption{Absolute performance between FPGA and ARM}
\label{fig:absoluteimprovement}
\end{figure}

The ARM processor implementation, on the other hand, which relies on the conventional design of a CPU integrated into the Xilinx ZCU102 platform, is unable to match the performance of the FPGA implementation in terms of the total time required to complete the evolutionary process.

Figure 1 presents the comparison between the FPGA and ARM processor implementations for different time steps per oscillation values. The y-axis represents the total time required to complete the evaluations on a log scale. As can be observed, the FPGA implementation consistently outperforms the ARM processor implementation across all time steps per oscillation values. The difference in performance becomes more pronounced as the number of time steps per oscillation increases, highlighting the advantage of the FPGA's parallelism and reconfiguration speed.

Figure 2 illustrates a pound-for-pound comparison of the the most computationally burdensome evaluation task: 1000 Euler steps with an increasing population up to 1,000,000 circuits. A clear distinction arises that the FPGA becomes exponentially more performant with an increase in population size.

\begin{figure}[h]
\centering
\includegraphics[width=1.0\linewidth]{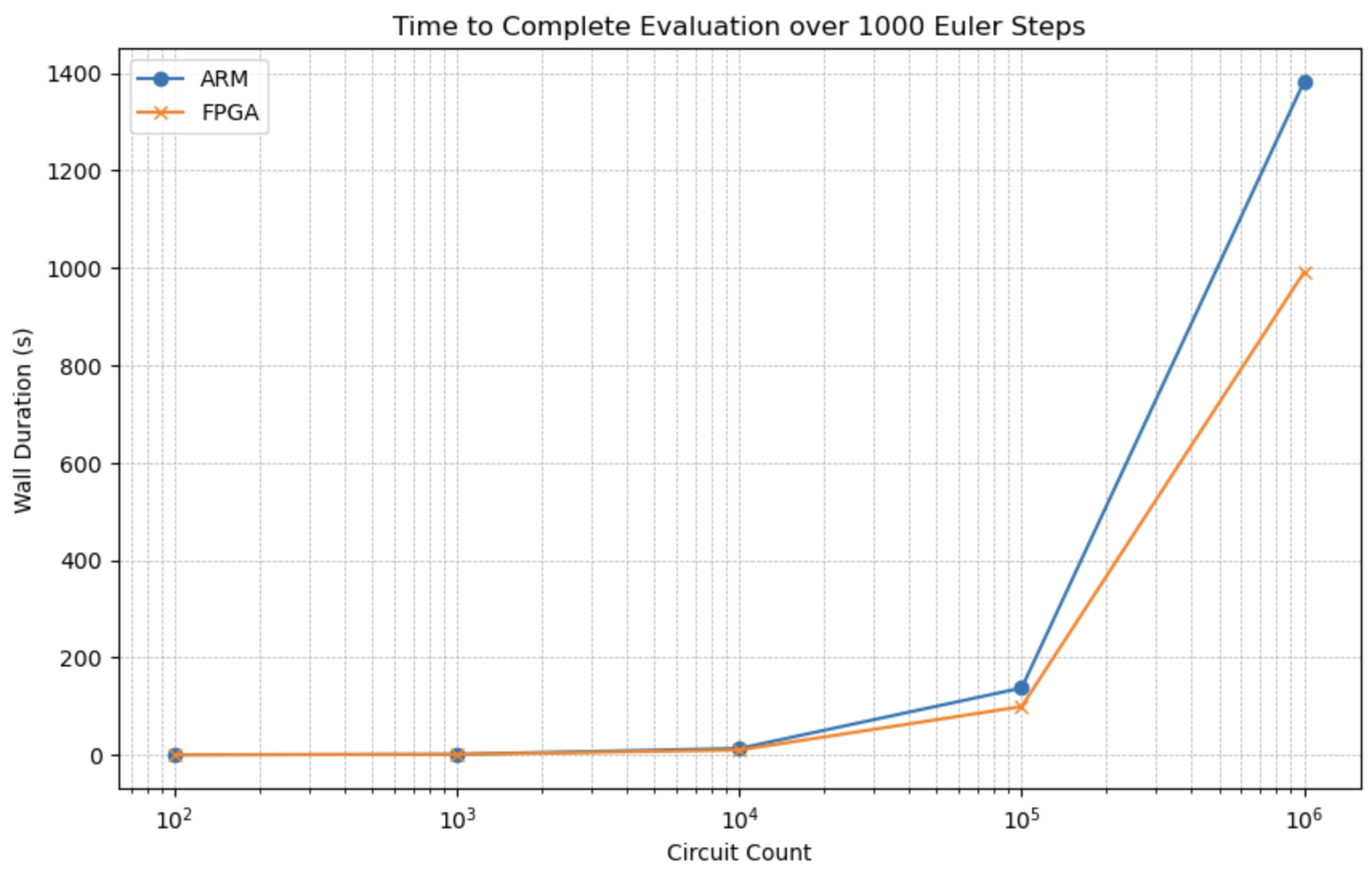}
\caption{Initial comparative performance between FPGA and ARM}
\label{fig:fabricimprovement}
\end{figure}

Lastly, Figure 3 highlights the percentage improvement of utilizing an FPGA over ARM as function of increasing integration period and population size. The FPGA hovers between 20-28\% faster than the ARM core alone.

\begin{figure}[h]
\centering
\includegraphics[width=1.0\linewidth]{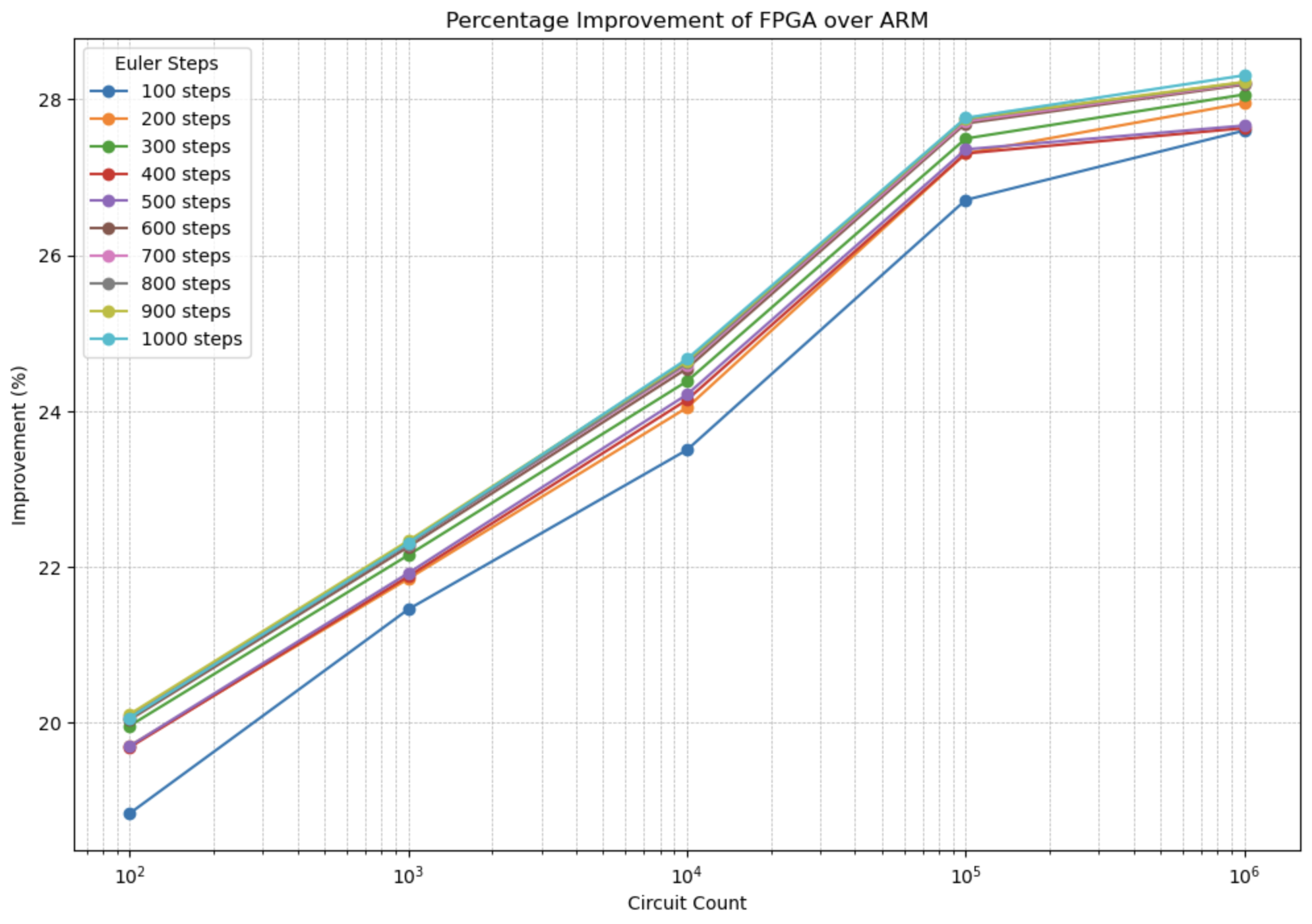}
\caption{Percentage Improvement of FPGA over ARM}
\label{fig:percentageimprovement}
\end{figure}

\subsection{Discussion}

The results of the comparison between the FPGA and ARM processor implementations reveal the significant advantages that FPGAs offer for neuroevolutionary algorithms involving CTRNNs. The FPGA's ability to evaluate a large number of CTRNNs in parallel and update the next generation of CTRNNs using dynamic and partial reconfiguration leads to a substantial reduction in the total time required to complete the evolutionary process.

In particular, the performance advantage of the FPGA implementation becomes more evident as the number of time steps per oscillation increases. This is likely due to the fact that the FPGA's parallelism and reconfigurability become more beneficial as the workload increases. In contrast, the ARM processor's performance is constrained by the limited number of available cores and the efficiency of the algorithm's implementation on the processor. As the number of time steps per oscillation increases, the ARM processor's performance does not scale as effectively as the FPGA's, resulting in a larger performance gap between the two implementations.

\section{Conclusion}
Though additional work is necessary, the results of this initial study have demonstrated the significant advantages of implementing continuous-time recurrent neural networks (CTRNNs) on field-programmable gate arrays (FPGAs) and using dynamic and partial reconfiguration (DPR) to expedite the evaluation period in neuroevolutionary algorithms. FPGAs offer substantial benefits in terms of parallelism, reconfigurability, and performance, allowing for the efficient execution of neuroevolutionary tasks and the rapid exploration of large solution spaces.

In comparison to the ARM processor-based implementation, the FPGA implementation consistently outperformed across different time steps per oscillation values. This can be attributed to the FPGA's ability to evaluate a large number of CTRNNs in parallel, reducing the total time required to complete the evolutionary process. Additionally, the use of DPR enables the updating of the next generation of CTRNNs without the need for a full reconfiguration of the FPGA, further reducing the time required for each generation.

The success of implementing CTRNNs on FPGAs and using DPR to expedite the evaluation period highlights the potential of FPGAs as a viable alternative to graphics processing units (GPUs) for deep learning tasks. GPUs have been the de facto choice for deep learning due to their massive parallelism, high computational throughput, and ease of use. However, FPGAs offer several advantages over GPUs, such as lower power consumption, higher performance per watt, and the ability to reconfigure hardware on-the-fly to adapt to changing requirements. As demonstrated in this study, these advantages can be leveraged to accelerate the neuroevolution of CTRNNs and reduce the overall time required for the optimization process.

Despite these advantages, FPGAs have not yet achieved the same level of popularity and widespread adoption as GPUs for deep learning tasks. One of the main barriers to FPGA adoption in this domain is the ease of use. Developing and deploying deep learning models on FPGAs typically involves a more complex design flow, requiring expertise in hardware description languages, digital logic design, and optimization techniques. This can be a deterrent for researchers and practitioners who are accustomed to the simplicity and accessibility of deep learning frameworks and libraries for GPUs.

To foster the broader adoption of FPGAs for deep learning tasks, future work should focus on making FPGAs as easy to use with CTRNNs as GPUs currently are. This can be achieved by developing high-level programming frameworks and libraries that abstract away the complexity of FPGA design and allow researchers and practitioners to easily implement and deploy CTRNNs on FPGAs. Such frameworks should provide a similar user experience to existing deep learning frameworks for GPUs, allowing users to define, train, and evaluate CTRNNs with minimal effort and without the need for extensive hardware design expertise.

Moreover, future work should also explore techniques for automating the process of DPR for CTRNNs on FPGAs, simplifying the management of partial bitstreams and reducing the overhead associated with reconfiguration. By automating these processes, researchers can focus on the design and optimization of CTRNNs, rather than the low-level details of DPR and FPGA management.

Another direction for future work is the development of optimized hardware architectures for CTRNNs on FPGAs, leveraging the reconfigurability and flexibility of FPGAs to tailor the hardware specifically for CTRNNs. By designing hardware architectures that are optimized for CTRNNs, it may be possible to further improve the performance, power efficiency, and scalability of FPGA-based implementations.

In conclusion, the success of implementing CTRNNs on FPGAs and using DPR to expedite the evaluation period showcases the potential of FPGAs as a powerful and efficient platform for neuroevolutionary tasks. However, to unlock the full potential of FPGAs in this domain, future work must focus on improving the ease of use, developing high-level programming frameworks, and optimizing hardware architectures specifically for CTRNNs. By addressing these challenges, FPGAs could become a widely adopted and accessible platform for deep learning tasks, offering a compelling alternative to GPUs.

\footnotesize
\bibliography{FPGA_neuro}
\bibliographystyle{apalike}
\end{document}